\documentclass[sigconf,nonacm]{acmart}
\AtBeginDocument{%
  }

\begin{document}
\title{Fairness is Not Flat: Geometric Phase Transitions Against Shortcut Learning}

\author{Nicolas Rodriguez-Alvarez}
\email{nicolas.rodalv@educa.jcyl.es}
\orcid{0009-0002-6804-386X}
\affiliation{%
  \institution{Instituto de Educaci\'on Secundaria Parquesol}
  \city{Valladolid}
  \state{Castile and Le\'on}
  \country{Spain}
}

\renewcommand{\shortauthors}{Rodriguez-Alvarez}

\begin{abstract}
Deep Neural Networks are highly susceptible to shortcut learning, frequently memorizing low-dimensional spurious correlations instead of underlying causal mechanisms. This phenomenon not only degrades out-of-distribution robustness but also induces severe demographic biases in sensitive applications. In this paper, we propose a geometric \textit{a priori} methodology to mitigate shortcut learning. By deploying a zero-hidden-layer ($N=1$) Topological Auditor, we mathematically isolate features that monopolize the gradient without human intervention. We empirically demonstrate a Capacity Phase Transition: once linear shortcuts are pruned, networks are forced to utilize higher geometric capacity ($N \geq 16$) to curve the decision boundary and learn ethical representations. Our approach outperforms L1 Regularization---which collapses into demographic bias---and operates at a fraction of the computational cost of post-hoc methods like Just Train Twice (JTT), successfully reducing counterfactual gender vulnerability from 21.18\% to 7.66\%.
\end{abstract}




\maketitle

\section{Introduction}
Biological intelligence is fundamentally economic. The human brain, acting as a 'cognitive miser' \cite{Fiske2013}, evolved to prioritize fast, low-effort heuristics over exhaustive logical reasoning—a duality famously described as System 1 and System 2 \cite{Vargas_Bianchi_2022}. This ancestral 'laziness' is not a flaw, but a survival mechanism designed to conserve metabolic resources. Paradoxically, modern Machine Learning (ML) models have inherited this same evolutionary pressure. Despite their digital substrate, neural networks exhibit a profound \textit{simplicity bias} \cite{NEURIPS2020_6cfe0e61}, naturally gravitating towards 'shortcuts' that offer the path of least resistance to minimize the loss function at the expense of causal understanding.

In tabular environments such as the Adult Census dataset, this algorithmic ``System 1'' manifests as a disproportionate reliance on direct data leakages. For instance, rather than learning the complex, multi-dimensional interactions between an individual's education, age, and hours worked, a standard deep model will exploit financial anomalies like \textit{Capital-Gain}. Much like a human cognitive miser who avoids complex mental arithmetic when a simple heuristic is readily available, the neural network collapses its hypothesis space onto the most accessible predictive feature, effectively bypassing true causal inference.

To expose and neutralize this behavior, we propose the \textit{A Priori Geometric Auditor}. By intentionally restricting the network's topological capacity to a single linear boundary ($N=1$ hidden neurons), we strip the model of its ability to ``reason'' geometrically. Forced into extreme cognitive starvation, the auditor is compelled to reveal the dataset's path of least resistance, deterministically isolating spurious shortcuts and data leakages before the primary, high-capacity model is ever trained. Our main contributions are threefold:

\begin{itemize}
    \item \textbf{The A Priori Geometric Auditor:} We introduce a strictly linear probing methodology ($N=1$) that successfully isolates topological shortcuts---such as the \textit{Capital-Gain} data leakage---without requiring human prior knowledge.
    \item \textbf{Capacity Phase Transition:} We empirically demonstrate that algorithmic fairness requires geometry. Once linear shortcuts are topologically pruned, the network is forced to utilize higher geometric capacity ($N \ge 16$) to curve the decision boundary and learn meritocratic representations.
    \item \textbf{Counterfactual Bias Mitigation:} Through stress testing, we prove that pruning low-dimensional shortcuts not only immunizes the model against direct data leakage but also drastically reduces secondary demographic biases, lowering gender vulnerability from 21.18\% to 7.66\%.
\end{itemize}

\section{Related Work and Baselines}
Addressing shortcut learning often involves post-hoc data reweighting or standard regularization techniques. However, these methods suffer from significant computational or ethical drawbacks when compared to our geometric \textit{a priori} approach.

\begin{figure}[h]
  \centering
  \includegraphics[width=\linewidth]{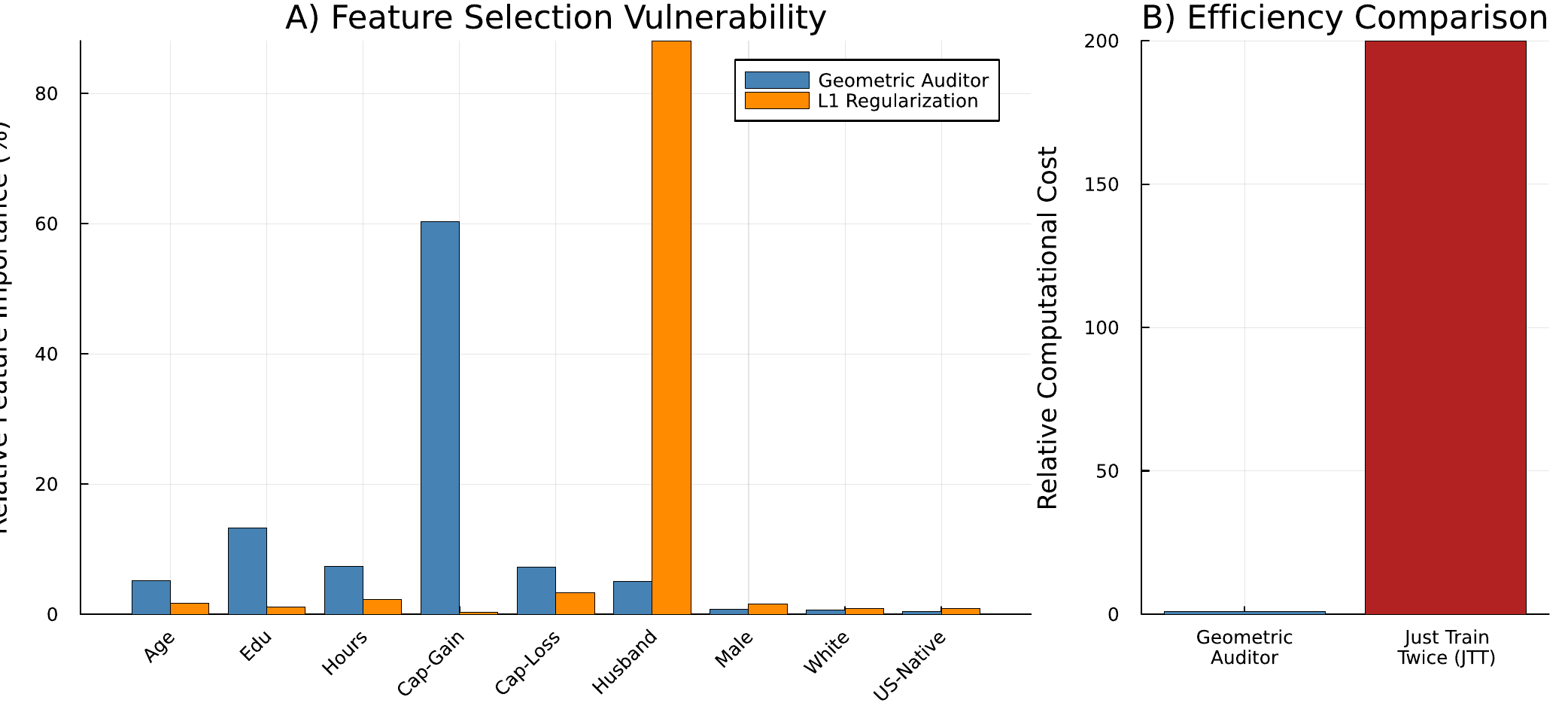}
  \caption{State-of-the-Art (SOTA) Comparison. (A) Relative feature importance: Auditor vs L1 Regularization. (B) Relative computational efficiency compared to JTT.}
  \label{fig:sota}
\end{figure}

\textbf{Standard Regularization (L1/Lasso):} L1 regularization is traditionally used to induce sparsity. However, it is entirely ``blind'' to the ethical or causal value of the features. As demonstrated in our empirical baseline (Figure \ref{fig:sota}A), applying an L1 penalty ($\lambda = 0.05$) catastrophically failed to mitigate bias. To minimize the regularization loss, the L1 model aggressively pruned true causal features (such as \textit{Education} and \textit{Hours worked}) and ultimately collapsed the entire hypothesis space onto a single protected attribute: \textit{Husband}. By contrast, our Topological Auditor successfully isolates the financial data leakage (\textit{Capital-Gain}) while preserving ethical merit variables.

\textbf{Just Train Twice (JTT):} Recent state-of-the-art approaches like JTT \cite{liu2021justtraintwiceimproving} attempt to mitigate spurious correlations by training an initial empirical risk minimization (ERM) model, identifying misclassified examples, and then training a second robust model. While effective, JTT doubles the computational cost and carbon footprint by requiring the training of multiple deep networks. Our Geometric Auditor offers a proactive alternative: by probing the dataset with $N=1$ hidden neurons, we mathematically isolate topological shortcuts at a fraction of the computational cost \textit{before} the robust model is ever trained (Figure \ref{fig:sota}B).

\section{Methodology: The Geometric Auditor}
Our approach relies on the principle that spurious correlations --such as direct data leakages-- are fundamentally low-dimensional and linearly separable, whereas true causal relationships in human behavior demand higher geometric capacity to be approximated. We propose a three-phase methodology to proactively mitigate shortcut learning.
\subsection{Linear Probing and Shortcut Isolation}
Following the intuition of linear classifiers to understand network representations, we deploy a zero-hidden-layer network ($N=1$) directly on the input features. The prediction for a given instance $x \in \mathbb{R}^d$ is given by standard logistic regression:
\begin{equation}
\hat{y} = \sigma \left( \sum_{i=1}^{d} w_i x_i + b \right)
\end{equation}
Lacking the geometric memory provided by hidden non-linearities, the optimizer is forced to exploit the simplest linear pathways to minimize the Binary Cross-Entropy loss. Consequently, features acting as spurious shortcuts monopolize the gradient, resulting in disproportionately large absolute weights \cite{alain2018understandingintermediatelayersusing}.

\subsection{Topological Pruning Criterion}
Rather than relying on human domain knowledge to identify these leakages, we implement an automated statistical pruning mechanism. We define a pruning threshold $\tau$ based on the mean absolute magnitude of the network's weights:
\begin{equation}
\tau = 2 \times \frac{1}{d} \sum_{i=1}^{d} |w_i|
\end{equation}
Any feature $x_i$ whose associated weight satisfies $|w_i|>\tau$ is mathematically flagged as a topological shortcut and is subsequently pruned from the hypothesis space prior to training the primary model.

\subsection{Capacity Phase Transition}
Once the low-dimensional shortcuts are removed, the dataset becomes strictly non-linear. To learn the underlying ethical representation (e.g., combining age, education, and hours worked), we train models of varying capacities. The robust model utilizes a deep architecture with non-linear ReLU activations:
\begin{equation}
f(x) = \sigma \left( W_2 \cdot \max(0, W_1 \cdot x + b_1) + b_2 \right)
\end{equation}
We search for the critical number of hidden neurons ($N$) where the model successfully curves the decision boundary to overcome the topological pruning, an event we define as the Capacity Phase Transition \cite{Goodfellow-et-al-2016}.

\section{Experiments and Results}
We validate our geometric approach using both synthetic environments (XOR with an artificial shortcut) and real-world tabular data (the Adult Census Income dataset).
\subsection{Automated Shortcut Detection}
\begin{figure}[h]
  \centering
  \includegraphics[width=\linewidth]{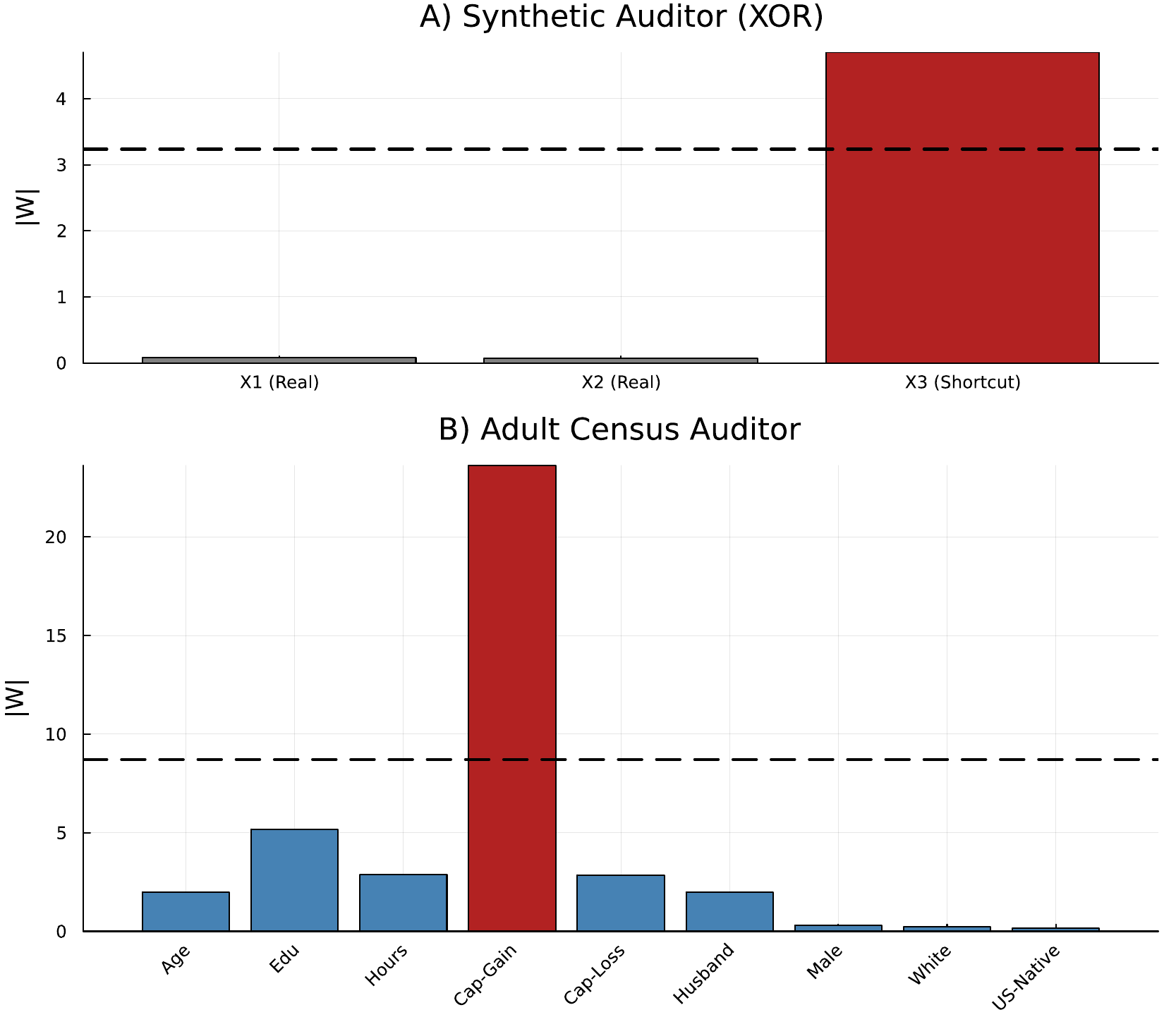}
  \caption{Automated Shortcut Detection using the Geometric Auditor ($N=1$).}
  \label{fig:auditor}
\end{figure}
As illustrated in Figure \ref{fig:auditor}, the linear auditor successfully identifies the dominant spurious correlations in both environments. In the Adult dataset, the Capital-Gain feature drastically exceeded the pruning threshold ($|w_{cap}|=23.63$, where $\tau=8.71$). This empirical finding confirms that standard models heavily rely on this specific financial variable as a direct shortcut to predict income, effectively ignoring educational and demographic merit.

\subsection{The Geometric Phase Transition}
\begin{figure}[h]
  \centering
  \includegraphics[width=\linewidth]{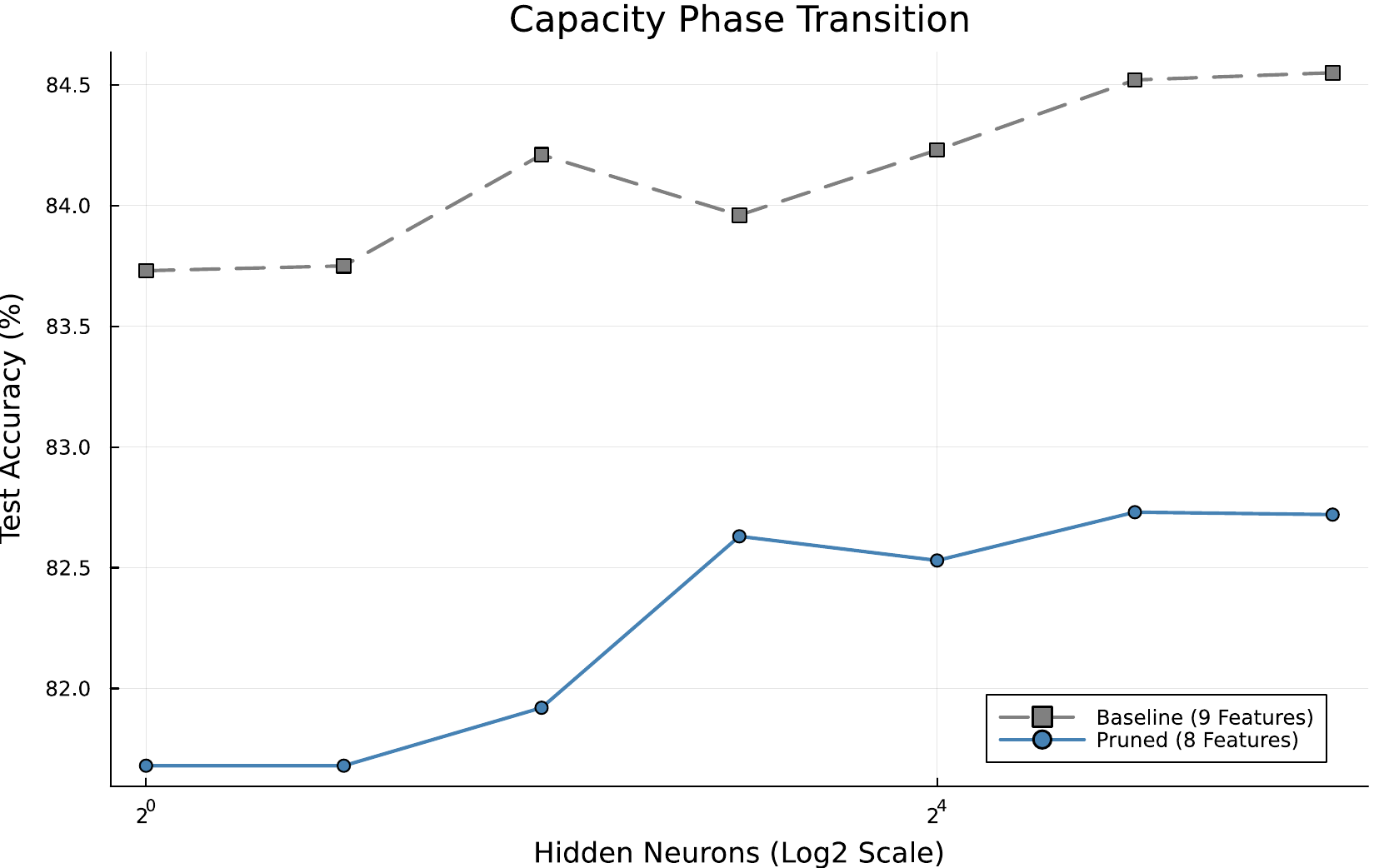}
  \caption{Phase transition between the biased and the pruned model.}
  \label{fig:capacity}
\end{figure}
Figure \ref{fig:capacity} demonstrates the necessity of geometric capacity for ethical learning. The unpruned baseline model achieves early convergence with minimal capacity due to its reliance on the linear shortcut. However, the pruned robust model exhibits a distinct topological phase transition. Deprived of the easy linear path, the network suffers from capacity starvation until it reaches $N \geq 16$ hidden neurons, at which point it successfully maps the complex, non-linear relationships of human effort, recovering an accuracy of $\sim 82.7\%$

\subsection{Counterfactual Stress Testing}
\begin{figure}[h]
  \centering
  \includegraphics[width=\linewidth]{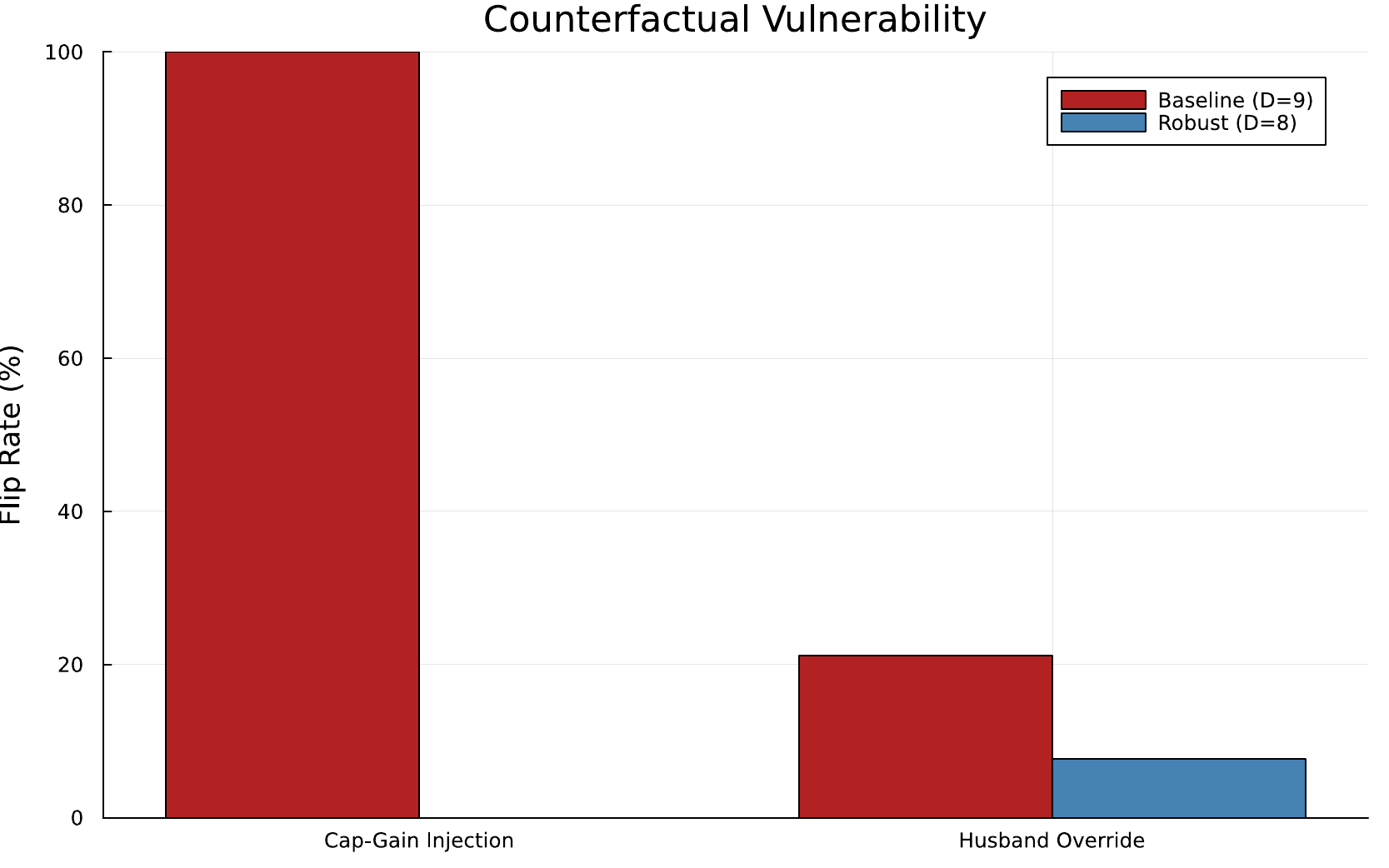}
  \caption{Counterfactual Vulnerability. Topological pruning eradicates data leakage (0\%) and significantly mitigates gender bias (Husband override).}
  \label{fig:vulnerability}
\end{figure}
Finally, we measure the models' robustness against data leakage and demographic bias using a counterfactual vulnerability framework \cite{NIPS2017_a486cd07}. We inject massive Capital-Gain values into low-income profiles and artificially override their marital status to Husband.
As shown in Figure \ref{fig:vulnerability}, the baseline deep model ($N=32$,$D=9$) exhibits a catastrophic $100\%$ flip rate when presented with fake capital gains, proving its absolute reliance on the shortcut. In contrast, our robust model ($N=32$,$D=8$) is structurally immune ($0\%$) to this leakage. More remarkably, the topological pruning induces a secondary ethical effect: by forcing the network to utilize its geometric capacity to learn meritocratic features, the model's vulnerability to gender bias (Husband override) dropped significantly from $21.18\%$ in the baseline to $7.66\%$ in the robust model.

\section{Conclusion}
In this work, we demonstrated that shortcut learning in deep neural networks is not merely a robustness issue, but a fundamental barrier to algorithmic fairness. By employing a zero-hidden-layer ($N=1$) Geometric Auditor, we mathematically isolated low-dimensional data leakages without relying on human priors. Furthermore, we showed that ethically sound decision-making requires a topological phase transition---deprived of linear shortcuts, networks must utilize higher geometric capacity to map complex human effort. Our approach proactively neutralizes data leakage and significantly reduces secondary demographic biases at a fraction of the computational cost of traditional methods. Robustness and fairness are not computationally flat; they require forcing the model to engage its full geometric capacity.

\begin{acks}
The author thanks Fernando Rodriguez-Merino (University of Valladolid) for valuable discussions on the research direction and for feedback on an early draft of this manuscript.
\end{acks}

\bibliographystyle{ACM-Reference-Format}
\bibliography{biblio}

\end{document}